# A ROBUST IRIS RECOGNITION METHOD ON ADVERSE CONDITIONS

Maryam Soltanali Khalili[1] and Hamed Sadjedi[2]

[1]Department of Electronic Engineering, Shahed University, Tehran, Iran
[1]Department of Electronic Engineering, Eslamshahr Azad University, Eslamshahr, Iran
[2] Department of Electronic Engineering, Shahed University, Tehran, Iran

## ABSTRACT

*As a stable biometric system, iris has recently attracted great attention among the researchers. However, research is still needed to provide appropriate solutions to ensure the resistance of the system against error factors. The present study has tried to apply a mask to the image so that the unexpected factors affecting the location of the iris can be removed. So, pupil localization will be faster and robust. Then to locate the exact location of the iris, a simple stage of boundary displacement due to the Canny edge detector has been applied. Then, with searching left and right IRIS edge point, outer radios of IRIS will be detect. Through the process of extracting the iris features, it has been sought to obtain the distinctive iris texture features by using a discrete stationary wavelets transform 2-D (DSWT2). Using DSWT2 tool and symlet 4 wavelet, distinctive features are extracted. To reduce the computational cost, the features obtained from the application of the wavelet have been investigated and a feature selection procedure, using similarity criteria, has been implemented. Finally, the iris matching has been performed using a semi-correlation criterion. The accuracy of the proposed method for localization on CASIA-v1, CASIA-v3 is 99.73%, 98.24% and 97.04%, respectively. The accuracy of the feature extraction proposed method for CASIA3 iris images database is 97.82%, which confirms the efficiency of the proposed method.*

## KEYWORDS

*IRIS recognition, pupil, edge detector, wavelet,*

## 1. INTRODUCTION

In recent years, application of biometric techniques to identify the individuals in various parts of society has been in the focus of attention. Fingerprints, palm print, face, voice, iris, hand geometry, and retina, which are invariant by natural factors (e.g., temperature, aging, disease, and climate variations), are well-known biometrics. In this context, much attention has been paid to the iris as a biometric system due to its intrinsic characteristics such as its life-time stability [1], uniqueness, reliability, taking image without physical contact (i.e. the ability to register optically without any contact), and the lowest error rate based on the statistical results [2]. The iris identification system works based on the precise location of the iris region in any eye image, extraction of the distinguishing features of the iris, and matching of the iris feature vectors using distance criteria. Different approaches have already been reported in the literature to determine the identity of people through their iris. Daugman [3] in 1993, for example, proposed the first efficient iris recognition system. He located the iris boundaries using a relatively time-consuming differential operator. He calculated the convolution of complex Gabor filters and iris image to extract the image features. Then he evaluated the complex map of phasors and generated a 2048-bit iris code so as to match the iris codes with Hamming distance criteria. Although the Gabor filter-based methods show a high degree of accuracy, they, nevertheless, require a long computational time. Wildes [4, 5] then used the gradient criterion and circular Hough transform to locate the iris. Besides, he proposed the application of Laplasian operator to extract the iris images features in 4 levels of accuracy (i.e., Laplasian with 4 different resolution levels) and used the normalized correlation coefficients for matching between the patterns of images. In reference [6] after the removal of the reflections on the image, Adaboost-cascade detector has been applied to extract the primary iris position. The border edge points have, then, been detected, and an elastic push-pull model has been used to determine the center and radius of the iris. Furthermore, a smoothing spline-based edge fitting scheme has been presented to find noncircular iris boundaries. In [7] and [8], the wavelet analysis has been used to determine the iris boundaries. Authors in [9] and [10] have proposed the use of morphological operators to locate the iris. In reference [11], a method has been proposed based on the solution of one-dimensional optimization problems using the incremental k-means (i.e., quantization) and Run-length encoding. This method was tested on the University of Bath Iris Database. Sunil Kumar Singla and Parul Sethi [12] have investigated the various problems which may arise in different modules of the iris biometric identification systems (e.g., sensor, pre-processing, feature extraction, and matching modules). Bimi Jain, et al. [13] have developed an iris recognition algorithm using Fast Fourier transform and moments. Authors in [14] using a combination of iris and signature biometrics and in [18] using a combination of iris images and brain neurons have sought to determine the individuals identification so that the fault detection of the system can be reduced. In another paper [15], the iris images are mapped in Eigen-space and then the iris code signature is generated from different camera snapshots of the same eye to incorporate the tonal and lighting variations. In another study conducted by Nithyanandam et al. [16], the Canny Edge Detector and circular Hough Transform have been used to detect the iris boundaries. The localized iris region is normalized and a basic phase correlation approach is applied to the image. After that, the normalized iris region is convolved with 2D Gabor filter so that the iris features can be extracted. The Hamming distance criterion is used to compare the iris codes. In the said study, the Reed-Solomon technique is employed to encrypt and decrypt the data. In other work [17], histogram equalization and wavelet techniques have been used to detect the iris. Cho, et al. [19] have calculated the correlation coefficients of the iris to compare the iris images. They have tested proposed algorithm on the iris images of the MMU1database. In reference [20], Perceptron neural network is proposed for iris recognition. Hollingsworth in [21] has used the information from the fragile bits to identify the iris. Authors in [22] have presented an identification system to be robust even when the images include obstructions, visual noise, and different levels of illuminations. In this work, decomposition of Harr wavelet up to 5 levels has been used to extract the iris features, and also the Hamming distance criteria has been applied for the matching of the feature vectors. The algorithm has been tested on images of CASIA and UBIRIS databases. In reference [23],the quality of the degradation iris images in visible light has been evaluated and the way to use the information about these damaged images has been explained. Sundaram, et al. [24] have employed the Hough transform to locate the iris but tried to reduce the searching complexity of Hough transform for the sake of less computational cost. Authors in [25] have aimed to present an algorithm for localizing the iris in non-cooperative environments. The inner boundary has been obtained using morphological dilation operator, Hough transform, and Otsu method. Besides, the outer boundary has been detected using vector machine classifier (SVM). This algorithm has been tested on UBIRIS database.

Different parts of an iris identification system are affected by various factors, such as lighting conditions, specular reflections, occlusion part of iris region by eyelashes and eyelids, rotating head, and contact lenses. Consequently, the accuracy of the identification system is reduced. All of the above-mentioned studies have sought to develop effective methods to address some of the problems involved. An efficient method for accurate classification (i.e., localization) of the iris is to use the edge detector [5], [6]. Application of the wavelet conversions to the iris feature extraction process is very effective in finding the characteristics of the iris texture [7], [8]. Moreover, the use of the matching absolute distance criteria will lead to a rapid identification system [3].

## 2. IRIS localization

Locating the pupil boundary is composed of three stages, pre-processing, location of the center and radius of pupil and post processing.

### 2.1. pre-processing

There are confusion factors in the basic eye images such as eyelashes, eyelids, contact lenses, dark or bright spots and specular reflections that are difficult to locate the exact iris. So applying an efficient pre-processing procedure to neutralize the effect of these factors is particular importance. Pixels of pupil region have the lowest brightness in eye images. So you can highlight pupil region and smooth confusion boundaries to locate the pupil boundary. According to equation 2-1, apply an Averaging filter results smoothing boundaries of the image.

$$I_{new}(i,j) = I(i,j) + \frac{1}{(2*m+1)(2*n+1)} \sum_{y=j-m}^{j+m} \sum_{x=j-n}^{i+n} I_{old} \qquad (2\text{-}1)$$

Combination of original image and the weighted mask of image is an image in which the pupil boundary remains and other elements are reduced as much as possible. Equation 1 smooth all borders consist of unwanted eye image borders and the pupil boundary. For this reason it is important how to influence (effect) the mask image on all components except the papillary boundary. To achieve this goal according to equations 2-2 and 2-3 using a weighted mask will be effective. The rate of change of each pixel is determined by their importance in the pupil area.

$$S_r(i) = \frac{1}{m} \sum_{i=1}^{m} I(i,:)$$

$$S_c(i) = \frac{1}{n} \sum_{i=1}^{n} I(:,i) \qquad (2\text{-}2)$$

$$I_w(i,j) = \frac{a}{2} \sum_{i=1}^{m} \sum_{j=1}^{n} (S_r(i) + S_c(j)) \quad 0 < a < 1 \qquad (2\text{-}3)$$

'$S_r$' and '$S_c$' are respectively the rows and columns mean of pixels, 'I' is the basic eye image, m and n are the pixel dimensions of the image, and $I_w$ is a weighted mask of the eye image. In addition, 'a' is a numerical constant which is known from experience.

According to results the most relevant (appropriate) value of 'a' equal (is considered) to 2% of the average intensity of pixels. An example of this process on an eye image is shown in Fig. 1.

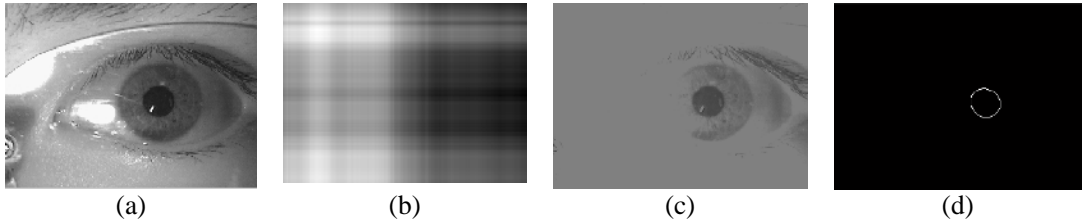

|  (a)  |  (b)  |  (c)  |  (d)  |

Figure 1: a) input image, b) weighted matrix mask c) Image obtained from the combined weight of the mask and the threshold applied to the input image D) Resultant image of applied the Canny edge detector on the image (c)

Then, as shown in Figure 1 - (c) pupil area can be distinguished is determined by applying a thresholding step.

$$I_n = I + I_W \tag{2-4}$$

$$I_t = (I_n < T) * I_n + T * (I_n > T) \quad 200 < T < 256 \tag{2-5}$$

'I' is the input image, '$I_w$' is the weighted matrix, '$I_t$' is the resultant image of the thresholding step. 'T' is the threshold level, of which the most appropriate value is 256 – as indicated by the test results.

As shown in figure 1-c by applying the proposed method, we were able to highlight pupil area than surrounding areas. Then the pupil boundary can be detected by applying the Canny edge detector. Canny edge detector can be used at this stage; with a threshold and the standard deviation to achieve the most suitable shape of the pupillary border. The result image of applying the Canny edge detector is shown Figure 1- d.

### 2.1.1. center and radius pupil

According to trigonometric theorem, perpendicular chord of circle pass through the center of the circle. So, central points can be obtained using the perpendicular chord passing through the boundary points. In this study, to reduce the computation cost only 5 points on the boundary are used. Average central points can be considered as the center of the pupil and the average distance of the pupil center to pupil boundary points can be considered as the radius pupil. The work is described in equations 2-6 to 2-8.

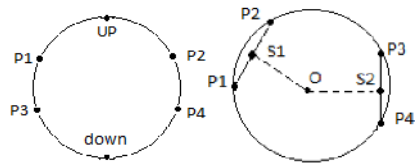

Figure 2: Find pupil boundary points and determine the center and radius of pupil

$$X_o = \frac{Y_{S2} - Y_{S1} - M_{S2o} * X_{S2} + M_{S1o} * X_{S1}}{M_{S1o} - M_{S2o}} \tag{2-6}$$

$$Y_o = M_{S1o}(X_o - X_{S1}) + Y_{S1} \tag{2-7}$$

$$R_p = \frac{1}{n} \sum_{n=1}^{a} O(X(n), Y(n)) \qquad (2\text{-}8)$$

'$X_O$' and '$Y_O$' are the coordinates of the pupil center, '$S_{1O}$' and '$S_{2O}$' are the perpendicular of '$P_1P_2$' and '$P_3P_4$' chords, '$S_{2O}$', '$M_{S1O}$' and '$M_{S2O}$' are the slopes of '$S_{1O}$' and '$S_{2O}$', $R_P$ is the radius of the pupil, and 'a' is the number of central points obtained from the intersection of any two perpendiculars.

Calculate the average has of the minimum complexity and low accuracy. Therefore, it is necessary to implement a post-processing step to increase the accuracy of the pupil center and radius values.

### 2.1.2. Post processing

In order to increase the accuracy of the values obtained for the radius and center of the pupil the pupil border should be moved toward the four directions and thus the most appropriate boundary location to obtain. To meet this goal, the boundary points are located in the four basic directions are used and the inner and outer boundary points are checked and if necessary, the border moved in the appropriate direction. For example, according to Figure 3 - (a) if the intensity of inner points at upper edge of the boundary in the input image is greater than the average intensity of the pupil region and intensity of the outer points of the bottom edge of boundary is lower than average intensity of the pupil area, pupil boundary is displaced downward. Different modes of transport boundary based on conditions are shown in Figure 3.

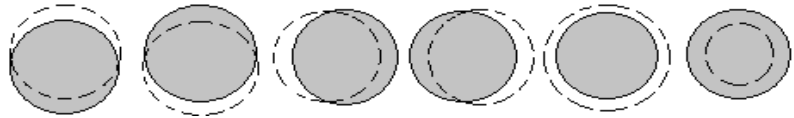

Figure 3: The correct location of the pupil boundary

In order to eliminate the adverse effects of bright spots of pupil area during correct location process, all pixels in the pupil area is replaced with the average intensity of pupil area.

In order to reduce computational cost, the average intensity of the pupil area is calculated in a square is inscribed with circle of the pupil (Figure 4.a).

$$sum = \sum_{i=X_o-R_p}^{X_o+R_p} \sum_{j=Y_o-R_p}^{Y_o+R_p} IM(i,j) \qquad (2\text{-}9)$$

$$IM(X,Y) = A * sum \quad 0 < A < 1$$
$$\sqrt{X^2 + Y^2} < R_p \qquad (2\text{-}10)$$

Where 'IM' is the input image and sum is the average intensity of the square area as shown in Figure 4.a. According to the test results, the most appropriate value for 'A' is equal to 0.02. 'A' acts as an index, and so it may also accept other values. The result is shown in Fig. 4-b.

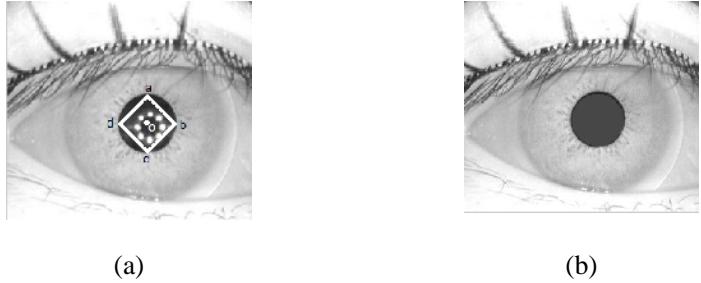

(a)                                     (b)

Figure 4: a) the averaging area b) replacement of the average intensity of pupil area instead of each pixel within the pupil area

## 2.2. Limbic boundary

The limbic and the pupil border have been considered as two concentric circles. So center and radius of pupil can be used to find the limbic boundary. First, Canny edge detector is applied on the input image to detect the limbic boundary (Figure 5.a). More images in the database, the upper half of the iris are occluded by eyelashes and the upper eyelid so limbic border cannot be available in this area. Hence, the bottom half of the iris is used to find points belongs to limbic border. The collarette boundary or boundaries of dark and light spots of the image are removed so that they do not create problems for finding the border points. The lower half of the iris between the two limbic and pupillary borders is clean. Due to the pupillary radius changes respect to limbic border in a given interval, the cleaned area is chosen. The result of the cleaning process is shown in Figure 5.b.

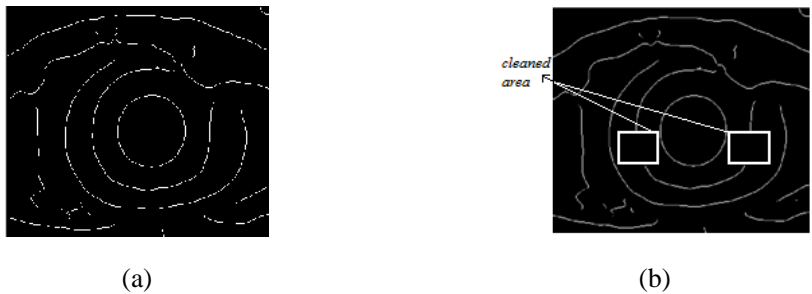

(a)                                           (b)

Figure 5: a) image after using edge detector b) remove the collarette border

Cleaned area obtains of the following equations based on the test results.

```
if (redius_pupil<33)
    a=ceil (2.85*redius_pupil)
elseif (redius_pupil<=35)
    a=ceil (2.65*redius_pupil)
elseif (redius_pupil<=36)
    a=ceil (2.69*redius_pupil)
elseif (redius_pupil<41)
    a=ceil (2.38*redius_pupil)
elseif (redius_pupil<47)
    a=ceil (2.15*redius_pupil)
elseif (redius_pupil<51)
    a=ceil (1.92*redius_pupil)
elseif (redius_pupil<55)
    a=ceil (1.7*redius_pupil)
else a=ceil (1.5*redius_pupil) end
```

Then, in the clean area and from the pupil boundary move to the sides and find some points belong to the limbic boundary. Among the finding points the values are far from the others will be removed and the average distance of boundary points from the center of the pupil is considered as limbic radius. Finally, a limbic radius value is controlled so that values more than maximum are replaced with the maximum value. This process is described below.

$$R_I=2*R_P \qquad 47<R_P<52, R_I>2.6*R_P$$
$$R_I=2.1*R_P \qquad 44<R_P<47, R_I>2.7*R_P$$
$$R_I=2.3*R_P \qquad 41<R_P<44, R_I>2.8*R_P$$
$$R_I=3*R_P \qquad 35<R_P<41, R_I>3 * R_P$$
$$R_I=2.6*R_P \qquad 35<R_P<41, R_I>3.3*R_P$$
$$R_I=3.4*R_P \qquad 31<R_P<35, R_I>3.4*R_P$$
$$R_I=3.6*R_P \qquad 23<R_P<31, R_I>3.5*R_P$$

'$R_P$' is the radius of the pupil and '$R_I$' is limbic radius.

## 3. Normalization

In an efficient iris identification system, after precise locating of the iris, the iris characteristics must be extracted using a safe method and then the images matching are matched through the application of these features. In Cartesian coordinates, the radius pupil value changes due to the variations in environment light or in the user's distance from the camera. This leads to a change in the size of the iris, which in turn causes the matching process to be difficult. Due to the fact that the information in the angle direction is more distinct than the radius information, the iris can be moved to the polar coordinates so as to investigate the iris features in these coordinates. Therefore, the located iris is partitioned to a rectangle with the dimensions 300 x 50 in polar coordinates. The length of the rectangular signifies polar variations, and the width of the rectangular is equal to the radial variations. Fig. 6 shows a normalized image.

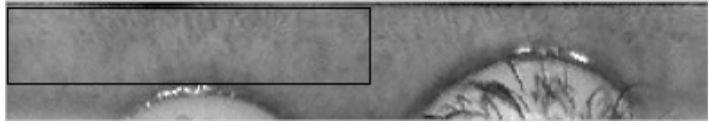

Figure 6: unwrapped Iris

Since in most images of the dataset, a large part of the upper half of the iris and a portion of its lower half are occluded by the eyelashes and eyelids, the area of the lower half of the iris is an appropriate ROI. In polar coordinates, the ROI is a rectangle with dimensions of 32 x 160, which is shown in Figure 6. Corresponding area of the rectangle in Cartesian coordinates is shown in Figure 7.

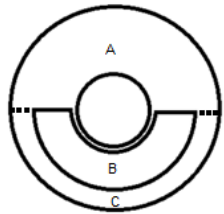

Figure 7: ROI region in Cartesian coordinates

In order to differentiate between the lower and the upper halves of the iris, the criterion expressed in Equation 3-1 to 3-3 can be utilized.

$$mean_A(i) = \sum_{j=1}^{n} F_i(j)$$

$$mean_B(i) = \sum_{j=1}^{n} F_i(j)$$

(3-1)

$$var_A(i) = \sum_{j=1}^{n} (F_i(j) - mean_A(i))^2$$

$$var_B(i) = \sum_{j=1}^{n} (F_i(j) - mean_B(i))^2$$

(3-2)

Where Fi (j) signifies the *i*th feature of the *j*th feature vector associated with one person, and 'mean' is the average of each of the features of all images related to one person. 'Var' is the variance within a class, and n is the number of the images belonging to the class. In other words, Fi (j) corresponds to the features belonging to a class and mean is the average of the features within the class.

(3-3)

$$Dis(A,B) = \frac{1}{n} \sum_{i=1}^{n} \frac{(\overline{mean_A(i)} - \overline{mean_B(i)})^2}{var_A(i) * var_B(i)}$$

'Dis' is the distance between the two classes.

Rapid changes in the texture of the iris are important to recognition. So the high-pass component and rapid changes should be strengthened in ROI region. For this purpose, first the low-pass components and slow changes should be extracted and removed from the ROI. This can be done by sliding an averaging window of size x over ROI area. Based on the test results, the appropriate value of x (between the values 4, 5, 6, 7, 8 and 9) has chosen to be 7. By subtracting the original image from the smoothed image, which acts as a background for the rapid changes, the pattern for these rapid changes in the iris texture can be obtained as shown in Figure 8 - (c). This process has been shown through equations 3-4 and 3-5.

(3-4)

$$IM_{background}(i,j) = \frac{1}{w^2} \sum_{m=i-w}^{i+w} \sum_{n=j-w}^{j+w} IM(m,n) \quad w = 7$$

$$I = IM - 0.9 * IM_{background}$$

(3-5)

'I' is the improved ROI region and w is the optimal averaging window.

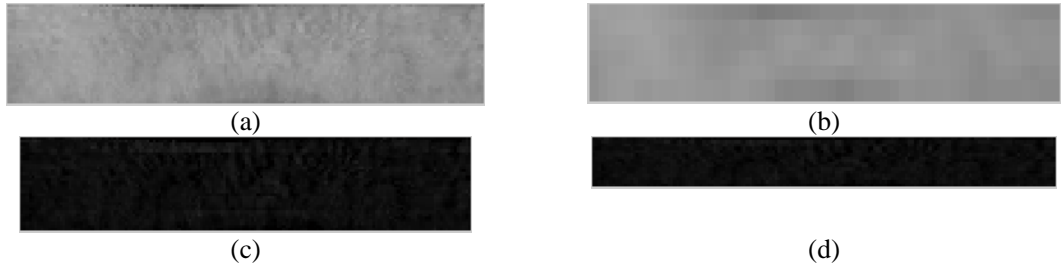

Figure 8: a) the ROI B) resultant image after using averaging window on the ROI c) the improved ROI d) the compressed ROI

In the next step, compression of the ROI area results in a decrease in the volume of the data and then of the calculations. As indicated by Equation 3-6, the method involved is to consider the average intensity of each two rows as a new row. Result of the implementation process is the ROI by the size 160 x 16 as shown in Fig. 8. D.

$$I_i(a,b) = \frac{1}{2}(I(i,j) + I(i+1,j)) \quad 1 < i < m \quad 1 < j < n$$

$$1 < a < \frac{m}{2} \quad b = j \tag{3-6}$$

# 4. Feature extraction

Discrete stationary wavelets transform 2-D (DSWT2) is one of the most powerful tools for solving the image processing problems. DSWT2 using a specific diagonal decomposition wavelet filters decomposes a multi-level analysis of both dimensions of the image. According to the flowchart in Fig. 9, approximation coefficients at level j is decomposed into 4 sub-images consisting of the approximate coefficients at level j +1 (ca), and partial coefficients in the horizontal (CH), vertical (CV), and diagonal (CD) directions.

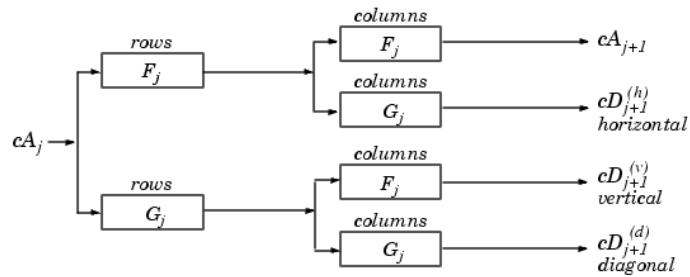

Figure 9: Flowchart of DSWT2, Rows box convolve entry rows and columns box convolve entry columns in filter X.

In the proposed paper, using DSWT2 and in the presence of the wavelet families including Symlet, Daubechies, Coiflet, Biorthogonal, and Reverse biorthogonal, the input image has been decomposed into sub-images at different levels. After investigation of the sub-images, the ones with the most distinguished features have been selected. Results of the sub-images analysis show that the approximation and vertical coefficients of the second level of the wavelet decomposition have distinctive features. Since a reduction in the size of the feature vectors results in a decrease in the computational cost, a stage of reduction for the feature vectors has also been implemented.

# 5. Matching

After the extraction of the features, an accuracy matching method with the least computational complexity is required to identify the users. Several matching methods are used to the iris identification systems. Some of these methods include the minimum distance methods based on the calculation of the Euclidean distance, hamming distance, absolute difference distance, and maximum similarity methods based on the calculation of correlation coefficient or the use of neural networks. Accurate correlation coefficient requires high computational cost. On the other hand, calculation of the absolute difference distance has a low accuracy and the least computational complexity. Since in an identification system, the rapidity of the algorithm – in addition to its accuracy – is also important, a matching method is required to satisfy both of these conditions. In the present study, it has been sought to combine the absolute difference distance and calculation of correlation coefficient to achieve an accurate fast matching method. Equations 5-1 to 5-4 illustrate this issue.

Calculate the correlation coefficient:

$$C(i,j) = \sum_{m=0}^{M_a-1} \sum_{n=0}^{N_a-1} A(m,n) * (B(m+i,n+j)) \quad 0 \le i < M_a + M_b - 1$$
$$0 \le j < N_a + N_b - 1$$

(5-1)

Matrices 'A' and 'B' are features of comparable images, '$M_a$', '$N_a$' and '$M_b$', '$N_b$' are dimensions of matrix 'A' and matrix 'B' respectively.

Calculate the absolute difference distance:

$$\sum_{i=1}^{n} |A(i) - B(i)|$$

(5-2)

Where matrices 'A' and 'B' are feature vectors of two images and 'i' is the number of features. Calculate the semi-correlation coefficient:

$$d(shift) = \frac{1}{M*N} \sum_{i=1}^{M} \sum_{j=1}^{N} |I_{fi}(i,j) - I_{fi}(i,j+shift)|$$

(5-3)

$$D_{min} = \min(d(shift)) \quad -5 < shift < 5$$

(5-4)

'$I_{fi}$' is matrix of the features of the input image; '$I_{fd}$' is matrix of the features of the database, and 'M' and 'N' are the size of features matrix.

According to equation 5-3, matrix of the features of the input image is compared with all stored features matrices and their shifts. Experimental results show that the minimum number of shift of features vector is 11. Therefore, it is no longer necessary to compare the features vector of input image with all shifts of stored features vectors.

Finally, to increase the accuracy of the decision about the identity of each user, K nearest neighbour (AKNN) is checked. Thus, $D_{mins}$ of any user are sorted, and at least three neighbours belonging to any class in the K first $D_{minS}$ determine the identity of the user. Otherwise, the minimum distance will be the result of k=1.

# 6. Result

The results of implementing the identification method by Matlab2009a on a computer with 2.6 GHZ processor and 4 GB RAM have been reported. Algorithm is tested on CASIA-v1 database consisting of 756 images of 108 persons, "interval iris images of CASIA3 database" containing 964 right iris images of 165 different people and 1318 left iris images of 185 persons. The image sizes are 280*320 for each eye. MMU1 database contains 211 left eye images of 42 persons and 222 right eye images of 45 persons. Interval iris images of CASIA3 database containing 964 right iris images are used to examine the features extraction and matching methods. Table 1 shows the accuracy of the proposed algorithm in iris localization.

Table 1: Results of the proposed algorithm is tested on CASIA database

| database | Number of users | Number of images | Accuracy of locating method (%) | Average of implementation time(sec) |
|---|---|---|---|---|
| CASIA1 | 108 | 754 | 99.73 | 0.221 |
| CASIA3-right eyes | 165 | 965 | 98.24 | 0.246 |
| CASIA3-left eyes | 185 | 1318 | 97.04 | 0.246 |
| MMU1-left eyes | 42 | 211 | 99.05 | 0.233 |
| MMU1-right eyes | 45 | 222 | 99.55 | 0.233 |

Accuracy of iris localization is 99.73%, 98.24% and 97.04%, respectively, and the average speed is 246ms. Images, in which the iris boundary is shifted and contain more than 5% of the other region or lose more than 5% of iris area, are accounted as the error images. All 964 right iris images of CASIA3 database are also used in the identification process. The accuracy of the identification algorithm is 97.82%. Results of applying all wavelets on database images are investigated to choose the appropriate wavelet.

At this stage, the absolute difference distance is used for matching images. Members of any family have similar behaviour. In view of this, only the best result of each family is shown in Table 2.

Table 2: iris recognition result of any wavelet family

| wavelet | Iris recognitions using ca,cv,ch and cd coefficient in level one(%) | Iris recognitions using ca,cv,ch and cd coefficient in level two(%) |
|---|---|---|
| Coif1 | 78.42 | 89.52 |
| Sym4 | 76.97 | 89.52 |
| Bior5.5 | 70.85 | 88.07 |
| Rbio2.2 | 79.14 | **91.28** |
| Db2 | 78 | 89.31 |

According to Table 1, the highest accuracy of iris recognition is equal to 91.28%, which has been obtained from two-level decomposition of the image and through a feature matrix with the dimensions up to 16*640. The large number of features increases the computations, which is a disadvantage for the identification system. Semi-correlation matching method can address this problem through reduction of the feature vector dimensions. The results in Table 2 are achieved using semi-correlation matching technique. Family members have similar behaviour, and so a representative sample has been used for each family. The dimensions of each of ca, cv, cd, ch matrices are 160 * 16.

Table 3: The accuracy of identification using a combination of different wavelet coefficients for different wavelet families

| Wavelet coefficients | Symlet4 (%) | Daubechies2 (%) | Coiflet1 (%) | Biorthogonal5.5 (%) | Reverse biorthogonal2.2 (%) |
|---|---|---|---|---|---|
| Ca2cv2ch2cd2 | 96.99 | 97.09 | 96.99 | 96.26 | 97.92 |
| Ca2cv2ch2 | 97.92 | 97.92 | 97.82 | 97.3 | 97.82 |
| Ca2cv2cd2 | 97.71 | 97.71 | 97.82 | 96.88 | 97.92 |
| Ca2cv2 | **98.03** | 97.82 | 97.92 | 96.99 | 97.2 |
| Ca2ch2 | 96.78 | 96.68 | 96.57 | 94.81 | 95.95 |
| Ca2cd2 | 97.2 | 97.2 | 97.4 | 95.43 | 96.99 |
| Ca2 | 96.99 | 96.88 | 96.78 | 94.5 | 94.5 |
| ca1cv1ch1cd1 | 91.39 | 91.8 | 92.22 | 87.76 | 92.63 |
| ca1cv1ch1 | 93.05 | 93.15 | 93.05 | 92.11 | 94.19 |
| ca1cv1cd1 | 93.67 | 93.77 | 93.88 | 89.62 | 93.98 |
| ca1cv1 | 95.85 | 95.64 | 95.54 | 94.6 | 96.78 |
| ca1ch1cd1 | 92.84 | 93.05 | 92.94 | 89.93 | 92.94 |
| ca1ch1 | 93.98 | 93.88 | 93.88 | 93.98 | 94.29 |
| ca1cd1 | 95.23 | 94.91 | 94.91 | 93.98 | 94.81 |
| ca1 | 97.51 | 97.3 | 97.3 | 97.9 | 97.61 |
| Ca2cd2ch2 | 96.47 | 96.37 | 96.57 | 95.33 | 96.78 |

As shown by Table 3, application of approximation and vertical coefficients of Symlet4 wavelet in second level results in the highest accuracy.

Correlation coefficients matching method increases the identification accuracy up to 99.37%. Therefore, it can be concluded that the shift factor reduces the accuracy of recognition. It should be noted that head rotation causes the shift of feature vector. Besides, 0.63% error in this approach is due to the error of localizing stage. Similarly, the semi-correlation matching technique is presented. Fig. 10 and Table 4 demonstrate the most optimal shift.

Table 4: Results of the different shifts of the ca, cv coefficients of sym4 wavelet in second level

| Number of shift of feature vector to left and right | Error (%) |
|---|---|
| 0 | 8.92 |
| 1 | 4.36 |
| 2 | 2.7 |
| 3 | 2.18 |
| 4 | 1.97 |
| 8 | 1.86 |
| 100 | 1.86 |

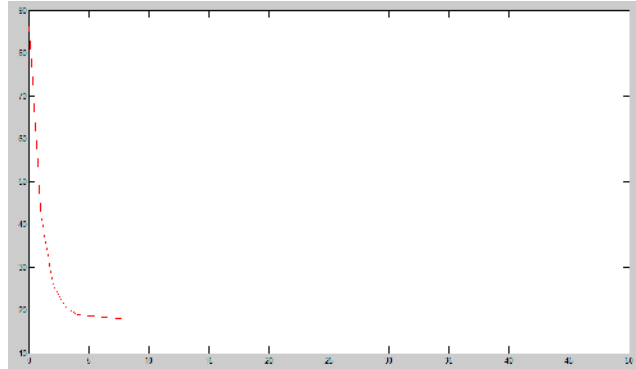

Figure 10: Graph of the number of shifts of the ca, cv coefficients of sym4 wavelet in second level

As is clear from Fig. 10, optimum number of feature vector shift is 4 time shift to both left and right directions. It is notable that shift in vertical direction has no effect on the identification. The aforementioned methods enhance the accuracy. However, as said before, the second most important factor in an efficient recognition system is high processing speed. Consequently, it is necessary to reduce the dimensions of the feature matrix of the order in a way that the accuracy of recognition is not reduced. As shown in Fig.11, there are duplicate data in vertical (radial) direction.

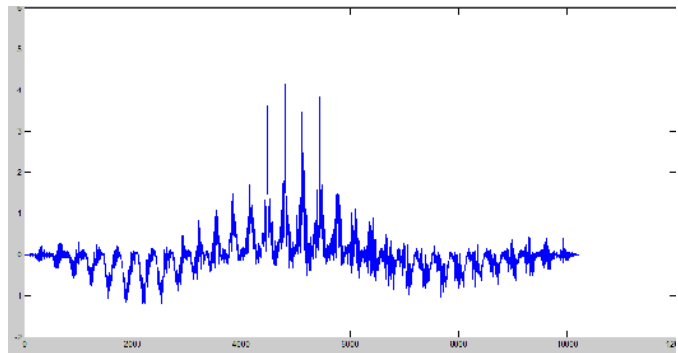

Figure 11: Calculation of the correlation coefficients between two feature vectors of a person

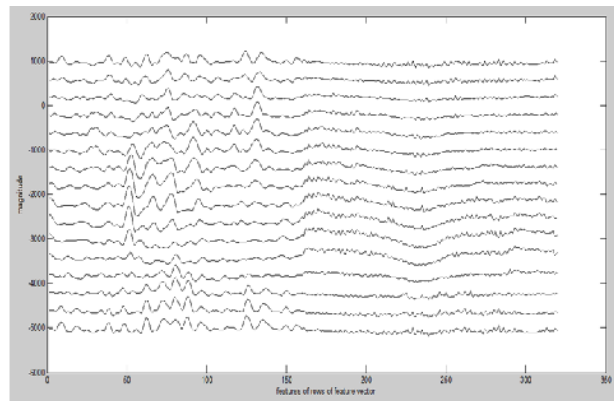

Figure 12: All rows of a person's feature matrix using Symlet 4 wavelet

45

Fig. 12 shows the data duplicate in rows of the features matrix. So, the first row of the features matrix is selected as the final feature vector. Notably, the feature matrix rows are shifted relative to each other to show the similarity of the matrix rows. Thus, the total number of features is reduced to 320. Moreover, due to the fact that the value of each feature is represented by two bits, 640 bits is equal to the length of the feature vector.

Table 5: The results of identification using the ca, cv coefficients of sym4 wavelet

| Feature vector | Length of feature vector (bit) | Number of shifts | Identification by AKNN, K=5, A=3 (%) | Normalization time (ms) | feature extract Time (ms) | matching time (ms) |
|---|---|---|---|---|---|---|
| cacv | 640 | 5 | 97.82 | 253 | 27.8 | 0.24 |

In order to show the effect of using two eyes on the identification of the users, the proposed method has been tested on interval iris image of CASIA3 database consisting of 600 both left and right eyes images. The test results are presented in Table 6.

Table 6: Results of identification using both of left and right eye images

| Number of images of each person | Error for using left eye images (%) | Error for using right eye images (%) | Error for using both eyes images (%) |
|---|---|---|---|
| Single registration | 9.1 | 10.2 | 5.8 |
| Three registration | 7 | 9.1 | 1.7 |
| Five registration | 2.5 | 1.8 | 0.66 |

As it can be seen in Table 6, with two eyes, the recognition accuracy would increase. Table 7 shows the result of changing the pre-processed conditions in the presence of ca, cv coefficients of Symlet4 wavelet.

Table 7: Change the Preprocessing conditions

| Preprocessing | Accuracy (%) |
|---|---|
| using the histogram equalize + apply compression | 51 |
| Averaging window by size of 7 * 7 + eliminates 90% of background brightness of the original image without compression | 41 |
| Averaging window by size of 7 * 7 + eliminates 90% of background brightness of the original image + using the histogram equalize + apply compression | 23 |
| Averaging window by size of 7 * 7 + eliminates 60% of background brightness of the original image + compression | 77 |
| Averaging window by size of 4 * 4 + eliminates 90% of background brightness of the original image + compression | 26 |
| Averaging window by size of 7 * 7 + eliminates 85% of background brightness of the original image + compression | 32 |
| Averaging window by size of 7 * 7 + eliminates 95% of background brightness of the original image + compression | 25 |
| Use the entire iris area as the ROI region | 93.25 |

Regarding the stated 'Dis' criteria and by calculation of sum (Dis) for comparison of Different information in three areas A, B, and C (as shown in Fig. 7), results of Table 8 have been obtained.

Table 8: scrutiny the non-distinct information of the non-ROI areas

| Discussed area | Non-distinct information (%) |
|---|---|
| B | 50 |
| C | 40 |

As shown in Fig.13, investigation of the enter class and entra class criteria can estimate efficiency of an identification system.

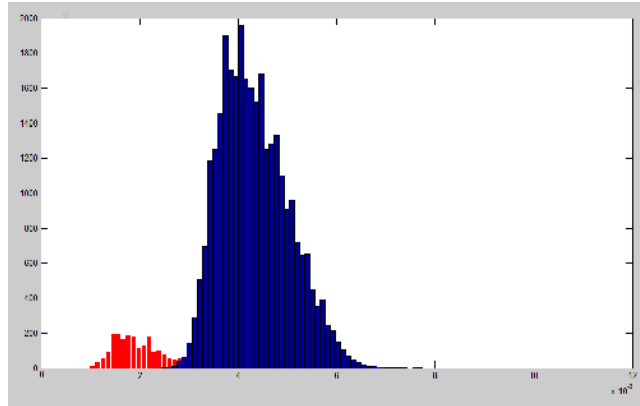

Fig.13: red diagram is enter class and blue diagram is entra class criteria

## 7. CONCLUSIONS

The present study has proposed an algorithm for the iris localization, which is robust against blurring images, specular reflections, disturbances of eyelids and eyelashes, wear glasses and unwanted edges. Since there are not disturbing reflections due to the application of glasses and wear glasses in the images of CASIA database, eye images of MMU1 database have been used. In the pre-processing of the ROI area by selecting the appropriate area mentioned above, non-distinct and redundant information has been removed. 'Dis' criteria indicates that there is 50% non-distinct information in the upper half of the iris and 40% in the lower part of the lower half of the iris. This is the reason why the ROI area has been selected. Averaging windows have also been used to highlight the high-pass components and the rapid changes as useful information of the iris area. Finally, the use of semi-correlation distance criteria as a fast accurate method has led to 97.82% accuracy of identification.

## REFERENCES

[1] John Daughman, "How Iris Recognition Works" IEEE Transactions on Circuits and Systems for Video Technology, Vol.14, No.1, January 2004.

[2] L.Flom and A.Safir, "Iris Recognition System. US. Patent 4641349, Patent and Trademark Office, Washington, DC, 1987.

[3] John Daugman, "High Confidence Visual Recognition of Persons by a Test of Statistical Independence", IEEE Transaction on Patern Analysis and Matchind Intelligence, Vol.15, No.11, November 1993.

[4] R. Wilde, J. Asmuth, G. Green, "A Machine-Vision System for Iris Recognition", Mach. Vis. Applic, 1996, 9, 1-8.

47

[5]  R. Wildes, "Iris Recognition: An Emerging Biometric Technology," Proc. IEEE, 1997, 85(9), 1348-1363.

[6]  Zhaofeng He, Tieniu Tan, Zhenan Sun, and Xianchao Qiu, "Toward Accurate and Fast Iris Segmentation for Iris Biometrics", IEEE Transactions on Pattern Analysis and Machine Intelligence, Vol.31, No.9, 1670-1684, September, 2009.

[7]  Cui, J., Wang, Y., Tan, T., Ma, L. and Sun, Z., "A Fast and Robust Iris Localization Method Based on Texture Segmentation", SPIE Defense and Security Symposium. Vol.5404, pp.401-408, 2004.

[8]  Shen, Y., Z., Zhang, M., J., Yue, J., W., and Ye, H., M.,"A New Iris Locating Algorithm", Proc. of Intl. Conf. on Artificial Reality and Telexistence – Workshops (ICAT'06), pp.438-441, 2006.

[9]  Miguel A. Luengo-Oroz a,d, Emmanuel Faure b,d, Jesús Angulo, "Robust Iris Segmentation on Uncalibrated Noisy Images Using Mathematical Morphology", Image & Vision Computing 28, 2009, 278-284.

[10] Feng Gui, Lin Qiwei, "Iris Localization Scheme Based on Morphology and Gaussian Filtering", Third International IEEE Conference on Signal-Image Technologies and Internet-Based System, 798-803, 2008.

[11] Nicolaie Popescu-Bodorin, "Exploring New Directions in Iris Recognition ",11th International Symposium on Symbolic and Numeric Algorithms for Scientific Computing Timisoara, Romania, September, 2009.

[12] Sunil Kumar Singla, and Parul Sethi, "Challenges at Different Stages of an Iris Based Biometric System", Songklanakarin Journal of Science and Technology (SJST), 2012.

[13] Bimi Jain, Dr.M.K.Gupta, Prof.JyotiBharti3, "Efficient Iris Recognition Algorithm Using Method of Moments", International Journal of Artificial Intelligence & Applications (IJAIA), Vol.3, No.5, September 2012.

[14] Serestina Viriri , Jules R. Tapamo, "Integrating Iris and Signature Traits for Personal Authentication Using User-Specific Weighting", Sensors - Open Access Journal,2012.

[15] Abhijit Das, Ranjan Parekh, "Iris Recognition Using a Scalar Based Template in Eigen-Space", International Journal of Computer Science and Telecommunications, Vol.3, Issue 5, May 2012.

[16] Nithyanandam.S, Gayathri.K.S, Priyadarshini P.L.K, "A New Iris Normalization Process for Recognition System with Cryptographic Techniques", International Journal of Computer Science Issues (IJCSI), Vol.8, Issue 4, No.1, July 2011.

[17] Khattab M. Ali Alheeti, "Biometric Iris Recognition Based on Hybrid Technique", International Journal on Soft Computing (IJSC), Vol.2, No.4, November 2011.

[18] Dr.T.Karthikeyan , B.Sabarigiri, "Enhancement of Multi-Modal Biometric Authentication Based on IRIS and Brain Neuro Image Coding", International Journal of Biometrics and Bioinformatics (IJBB), Vol.5, Issue 5, 2011.

[19] Eun suk Cho, Yvette Gelogo, Seok soo Kim, "Human Iris biometric Authentication using Statistical Correlation Coefficient", Journal of Security Engineering, 2011.

[20] M. Gopikrishnan, 2T. Santhanam, "Improved Biometric Recognition and Identification of Human Iris Patterns Using Neural Networks", Journal of Theoretical and Applied Information Technology, 2011.

[21] Karen P. Hollingsworth, Kevin W. Bowyer, and Patrick J. Flynn, "Improved Iris Recognition Through Fusion of Hamming Distance and Fragile Bit Distance", IEEE transactions on Pattern Analysis and Machine Intelligence, 2011.

[22] Vanaja Roselin.E.Chirchi , Dr.L.M.Waghmare , E.R.Chirchi, "Iris Biometric Recognition for Person Identification in Security Systems", International Journal of Computer Applications (0975 – 8887)V.24, No.9, June 2011.

[23] Hugo Proença, "Quality Assessment of Degraded Iris Images Acquired in the Visible Wavelength", IEEE Transactions on Information Forensics and Security, Vol.6, No.1, March 2011.

[24] R. Meenakshi Sundaram , Bibhas Chandra Dhara , Bhabatosh Chanda, "A fast Method for Iris Localization", Second International Conference on Emerging Applications of Infornation Technology, 2011.

[25] Tang Rongnian, Weng Shaojie, "Improving Iris Segmentation Performance via Borders Recognition", Fourth International Conference on Intelligent Computation Technology and Automation, 2011.

[26] CASIA Iris Image Database, http://www.sinobiometrics.com